\documentclass[letterpaper]{article} % DO NOT CHANGE THIS
\usepackage{aaai2026}  % DO NOT CHANGE THIS
\usepackage{times}  % DO NOT CHANGE THIS
\usepackage{helvet}  % DO NOT CHANGE THIS
\usepackage{courier}  % DO NOT CHANGE THIS
\usepackage[hyphens]{url}  % DO NOT CHANGE THIS
\usepackage{graphicx} % DO NOT CHANGE THIS
\usepackage{natbib}  % DO NOT CHANGE THIS AND DO NOT ADD ANY OPTIONS TO IT
\usepackage{caption} % DO NOT CHANGE THIS AND DO NOT ADD ANY OPTIONS TO IT
\usepackage{algorithm}
\usepackage{algorithmic}
\usepackage{amsmath}
\usepackage{booktabs}
\usepackage{multirow}

\usepackage{newfloat}
\usepackage{listings}
\DeclareCaptionStyle{ruled}{labelfont=normalfont,labelsep=colon,strut=off} % DO NOT CHANGE THIS
\floatstyle{ruled}
\newfloat{listing}{tb}{lst}{}
\floatname{listing}{Listing}
\title{DCA-LUT: Deep Chromatic Alignment with 5D LUT for \\ Purple Fringing Removal}
\author{
    Jialang Lu\equalcontrib\textsuperscript{\rm 1},
    Shuning Sun\equalcontrib\textsuperscript{\rm 2},
    Pu Wang\textsuperscript{\rm 3},
    Chen Wu\textsuperscript{\rm 4},
    Feng Gao\textsuperscript{\rm 5},
    Lina Gong\textsuperscript{\rm 6},
    Dianjie Lu\textsuperscript{\rm 7},
    Guijuan Zhang\textsuperscript{\rm 7},
    Zhuoran Zheng\textsuperscript{\rm 8}\thanks{Corresponding author.}
}
\affiliations{
    \textsuperscript{\rm 1}Hubei University\\
    \textsuperscript{\rm 2}University of the Chinese Academy of Sciences\\
    \textsuperscript{\rm 3}Shandong University\\
    \textsuperscript{\rm 4}National University of Defense Technology\\
    \textsuperscript{\rm 5}Ocean Unversity of China\\
    \textsuperscript{\rm 6}Nanjing University of Aeronautics and Astronautics\\
    \textsuperscript{\rm 7}Shandong Normal University\\
    \textsuperscript{\rm 8}Qilu University of Technology\\
    lujialang@stu.hubu.edu.cn, sunshuning23@mails.ucas.ac.cn,
    wangou@mail.sdu.edu.cn, wuchen5X@mail.ustc.edu.cn, 
    zhangzhihua@sdu.edu.cn, yz945@cornell.edu, gaofeng@ouc.edu.cn, gonglina@nuaa.edu.cn\{ludianjie, zhangguijuan\}@sdnu.edu.cn,
    zhengzr@njust.edu.cn
}

\usepackage{bibentry}
\begin{document}

\maketitle

\begin{figure*}[t]
\centering
\includegraphics[width=0.95\textwidth, height=0.3\textheight]{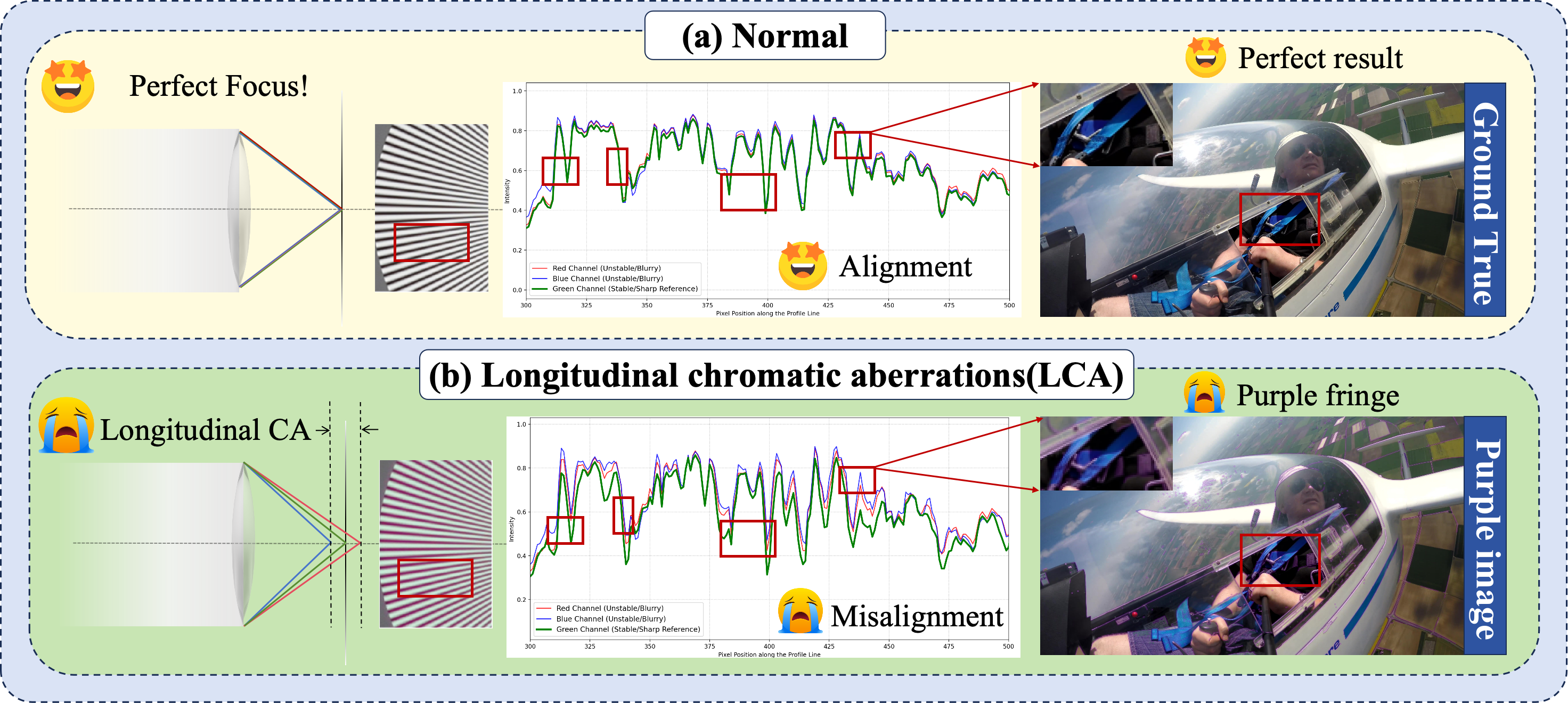}
\caption{This figure illustrates the physical root of Longitudinal Chromatic Aberration (LCA). 
  \textbf{(a)} Normal Case: All color channels are perfectly aligned, resulting in a sharp, clean image. 
  \textbf{(b)} Case with LCA: Different colors focus at different depths, causing red/blue shift relative to green, resulting in purple fringing.}
\label{fig: first}
\end{figure*}

\begin{abstract}
Purple fringing, a persistent artifact caused by Longitudinal Chromatic Aberration (LCA) in camera lenses, has long degraded the clarity and realism of digital imaging. Traditional solutions rely on complex and expensive apochromatic (APO) lens hardware and the extraction of handcrafted features, ignoring the data-driven approach. To fill this gap, we introduce DCA-LUT, the first deep learning framework for purple fringing removal. Inspired by the physical root of the problem, the spatial misalignment of RGB color channels due to lens dispersion, we introduce a novel Chromatic-Aware Coordinate Transformation (CA-CT) module, learning an image-adaptive color space to decouple and isolate fringing into a dedicated dimension. This targeted separation allows the network to learn a precise ``purple fringe channel", which then guides the accurate restoration of the luminance channel. The final color correction is performed by a learned 5D Look-Up Table (5D LUT), enabling efficient and powerful% non-linear 
color mapping. To enable robust training and fair evaluation, we constructed a large-scale synthetic purple fringing dataset (PF-Synth). Extensive experiments in synthetic and real-world datasets demonstrate that our method achieves state-of-the-art performance in purple fringing removal.
\end{abstract}


\section{Introduction}
In the pursuit of photorealism, digital imaging has become an indispensable technology. However, the quality of images is often compromised by optical artifacts inherent in lens systems. Among these, purple fringing is a persistent and visually disruptive artifact, manifesting as spurious purple or magenta halos around high-contrast edges. The physical root of this phenomenon is Longitudinal Chromatic Aberration (LCA)~\cite{10.1017/S155192950005481XLCA}, a simple camera lens refracts light of different wavelengths at slightly different angles, causing the focal planes of the primary colors (red, green, and blue) to separate~\cite{Volkova:19LCA, inproceedingsLCA}, illustrating in the Fig.~1. This spatial misalignment results in the characteristic fringing that degrades image sharpness and color fidelity. While hardware solutions like expensive apochromatic (APO) lenses can mitigate this issue~\cite{ibrahimdesign,articlelast}, they are often impractical for ubiquitous devices like smartphones. Consequently, developing effective computational methods for purple fringing removal remains a critical area of research.

\begin{figure}[h!]
    \centering
    \includegraphics[width=0.45\textwidth]{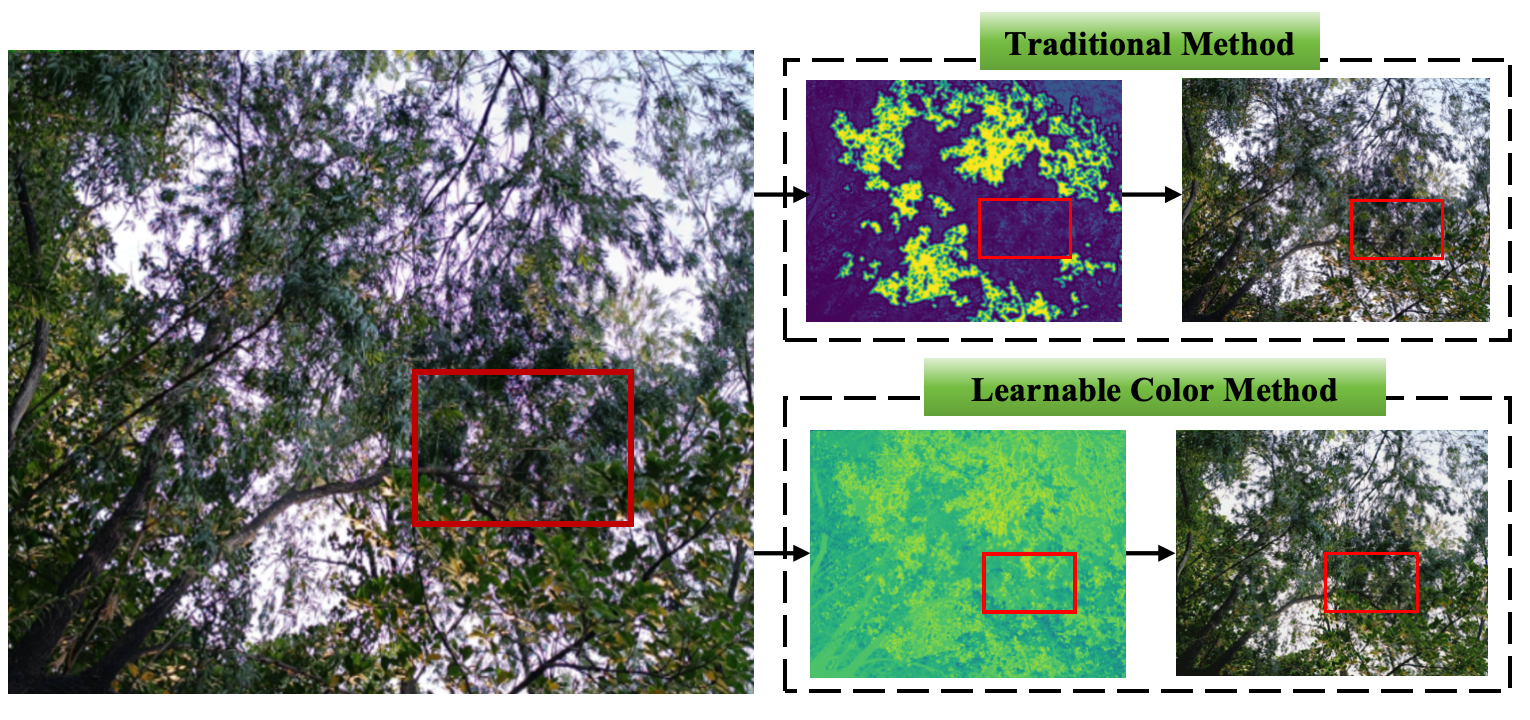}
    \caption{Comparison of fringe detection between traditional heuristic methods~\cite{article} and our learned CA-CT module. Our method accurately isolates the fringe artifact while preserving legitimate purple content.}
    \label{fig:cact_comparison}
    % \vspace{-0.6cm}
\end{figure}

Early computational approaches to this problem relied on handcrafted features to identify and correct purple fringing~\cite{kim2008automatic}. This traditional approach begins by identifying high-contrast edges where fringing is most common. Subsequently, it desaturates the specific purple chromaticity, which is characterized by red and blue channel values being notably higher than the green one~\cite{windowbase}. However, this rigid, rule-based logic is inherently unreliable. As shown in Fig.~\ref{fig:cact_comparison}, it often mistakes legitimate purple objects for artifacts or fails to completely remove the fringe, leaving behind residual color casts~\cite{kim2008automatic,bothpurlple}. Furthermore, their performance is often inconsistent across different lens types and lighting conditions, limiting their general applicability.

Neural networks powered by deep learning have led to significant advances in computational image restoration tasks such as denoising, deblurring, and super-resolution~\cite{JIANG2025103013denoise, zamir2022restormer,liang2021swinir}. This success has inspired initial efforts to apply deep learning to correct general optical aberrations, including chromatic aberration~\cite{maman2023achromatic}. 
Concurrently, learnable Look-Up Tables (LUTs) have emerged as a highly efficient method for performing complex, non-linear color transformations in image enhancement~\cite{zeng2020learning}. However, their application has been predominantly limited to global tasks like color grading, rather than precise, local artifact correction. We identify a critical opportunity to adapt this powerful and efficient technology to tackle the purple fringing problem directly.

Motivated by the physics of LCA, we introduce DCA-LUT to fill this gap, a framework that first decouples the fringe artifact into a dedicated channel, then uses that channel's geometric gradient to guide a direction-aware 5D LUT for precise, targeted correction. The main contributions of our work can be summarized as follows:
\begin{itemize}
\item We propose a Chromatic-Aware Coordinate Transformation (CA-CT) module inspired by the physics of LCA. This module learns to map the misaligned RGB channels into a new space, cleanly separating the purple fringe aberration into a dedicated channel.
\item We design a high-dimensional, gradient-guided 5D LUT module for chromatic aberration correction. It performs precise and efficient non-linear color correction. This design is targeted for purple fringing removal, where directional fringe cues are crucial for accurate correction.
\item To facilitate robust training and standardized evaluation, we construct \textbf{PF-Synth}, a large-scale synthetic dataset for purple fringing. We also propose the \textbf{Edge Chromatic Aberration Score (ECAS)}, a task-specific metric that quantifies residual color fringing along high-contrast edges, providing more accurate evaluation than PSNR and SSIM. Extensive experiments demonstrate that DCA-LUT achieves state-of-the-art performance on both synthetic and real-world images.
\end{itemize}

% To facilitate robust training and standardized evaluation, we have constructed PF-Synth, a new large-scale synthetic dataset for purple fringing, and extensive experiments show that DCA-LUT achieves state-of-the-art performance on both synthetic and real-world images.

\section{Related work.}
\textbf{Chromatic Aberration Correction.} Computational approaches to purple fringing correction have evolved from heuristic-based models to general deep learning frameworks. Early methods focused on detecting fringe regions using handcrafted features, such as edge information and color gradients, and then applied local corrections like color channel shifting or targeted desaturation~\cite{4270239, article}. However, as shown in Fig.~\ref{fig:cact_comparison}, these traditional techniques often struggle to generalize across diverse scenes and can introduce visual artifacts. With the rise of deep learning, Convolutional Neural Networks have been successfully applied to correct a broad range of optical aberrations by treating the problem as a blind deconvolution or geometric warping task~\cite{schuler2015learningcnn, Li_2019_CVPRblind, li2021universal}. While powerful, these general-purpose networks typically lack data-driven and physical inductive bias for the unique channel-specific defocus that characterizes Longitudinal Chromatic Aberration (LCA), failing to specifically model the root cause of purple fringing. Our work addresses this gap by introducing a framework specifically engineered to isolate and correct purple fringing based on its physical properties.
\begin{figure*}[h]
    \centering
    \includegraphics[width=\textwidth, height=0.4\textheight]{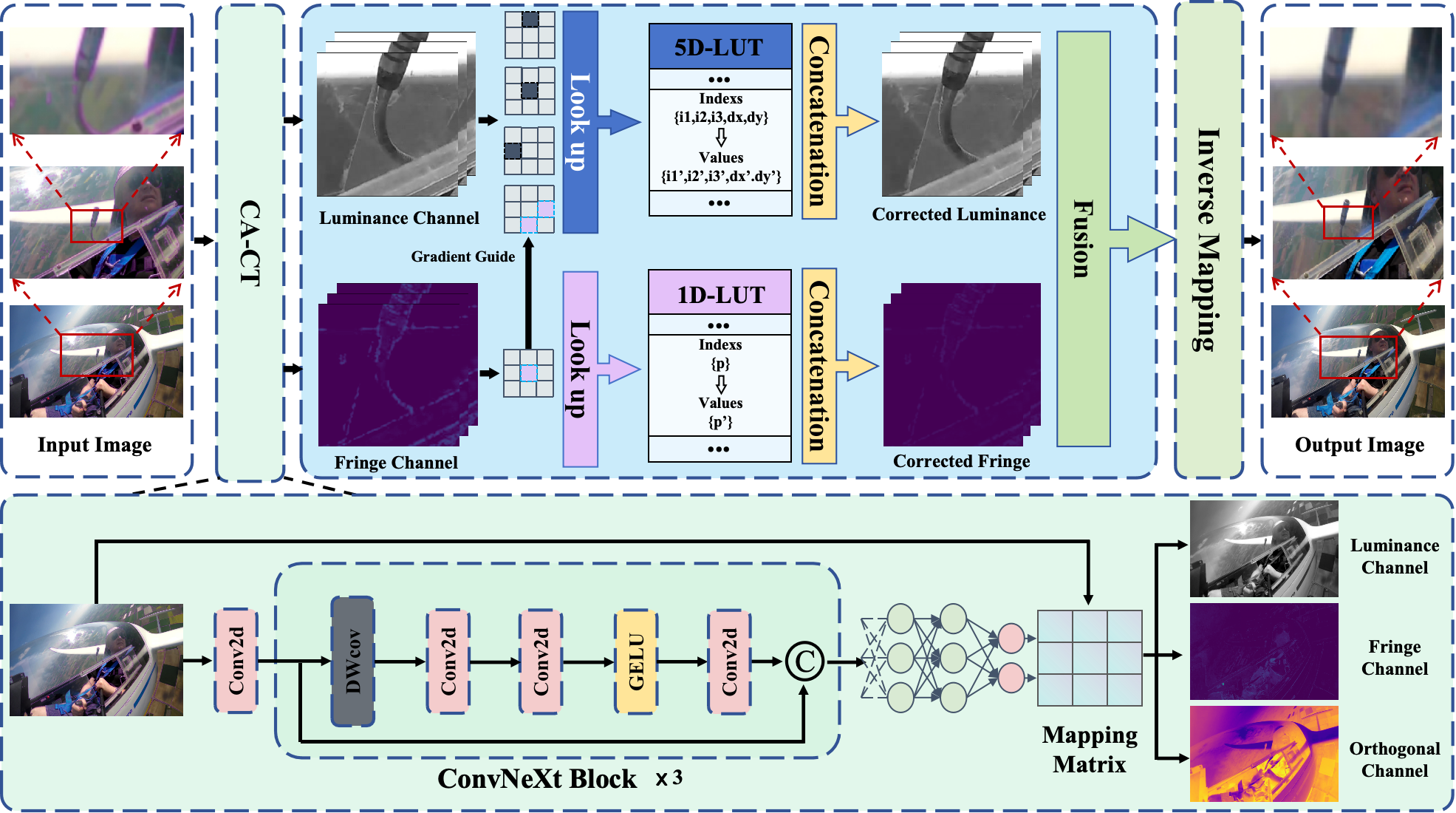}
    \caption{\textbf{The overall pipeline of DCA-LUT framework.} 
    \textbf{(a) Chromatic-Aware Transformation (CA-CT).} The input image is first processed by a ConvNeXt-based encoder to map into our learned Chromatic Aberration Space, decoupling the image into Luminance Channel ($C_{lum}$), Fringe Channel ($C_{fringe}$), and an Orthogonal Channel. 
    \textbf{(b) Direction-Aware Correction:} A dual-LUT correction branch, where the luminance channel is corrected by Direction-Aware 5D-LUT, utilizing the geometric gradients from the fringe channel to guide the luminance correction. Concurrently, the fringe channel is refined by a 1D-LUT. 
    %\textbf{(3) Fusion and Inverse Transformation:} 
    %The corrected luminance and fringe channels are fused and then transformed back to the standard RGB color space using the inverse matrix.
    }
    \label{fig:pipeline}
\end{figure*}

\noindent\textbf{Learnable Look-Up Tables for Image Enhancement.}
Integrating Look-Up Tables (LUTs) into deep learning pipelines has emerged as a highly efficient method for performing complex, non-linear color transformations. This was pioneered by learning a 3D LUT for real-time photo enhancement, a concept that was later extended to improve flexibility and reduce memory through innovations like separable structures and adaptive color space sampling~\cite{zeng2020learning, yang2022seplut, yang2022adaint}. To date, the application of learnable LUTs has been predominantly limited to global image enhancement tasks, such as color grading and style transfer~\cite{conde2024nilut, gong2025sa}. In contrast, we repurpose the LUT as a precise, local artifact correction module. We achieve this by introducing a novel direction-aware 5D LUT that utilizes cross-channel guidance, informing the geometric gradient of the isolated purple fringe channel into the LUT's coordinates to steer the final correction. Motivated by the physics of LCA, our direction-aware LUT directly counteracts the local color channel misalignment, enabling targeted, orientation-sensitive corrections that efficiently eliminate residual color shifts on resource-constrained devices.

\section{Methodology} 
\label{sec:method} Our proposed framework, DCA-LUT, is engineered to remove purple fringing by first transforming the input image into a problem-specific, learned color space. Within this space, the isolated luminance and fringe channels are independently corrected by LUT. The entire process is trained end-to-end, guided by a composite loss function that ensures both reconstruction fidelity and the physical plausibility of the learned transformations. The overall architecture is illustrated in Fig.~\ref{fig:pipeline}. 

\subsection{Chromatic-Aware Transformation} 
To isolate purple fringing, the first stage of our methodology generates image-specific parameters for transforming the input RGB image $\mathbf{I}_{RGB}$ into our proposed Chromatic Aberration Space (CAS). This is achieved via an image-adaptive $3\times3$ matrix $\mathbf{M}$, predicted by our Chromatic-Aware Coordinate Transformation (CA-CT) module. This module utilizes a lightweight ConvNeXt-based encoder to regress the nine elements of $\mathbf{M}$ from the global features of the input image. The transformation is a per-pixel matrix multiplication. For each pixel $p$ with color vector $\mathbf{I}_{RGB}(p)$, the corresponding pixel in the Chromatic Aberration Space is computed as:
\begin{equation} 
\mathbf{I}_{CAS}(p) = \mathbf{M} \cdot \mathbf{I}_{RGB}(p).
\end{equation} 
This transformed pixel $\mathbf{I}_{CAS}(p)$ is composed of a luminance channel ($C_{lum}$), a fringe channel ($C_{fringe}$), and an orthogonal channel ($C_{ortho}$). To ensure the stability and physical plausibility of this learned transformation, we introduce an \textbf{axis alignment loss}, ${L}_{align}$, which encourages the basis vectors of the CAS to be mutually orthogonal: \begin{equation} {L}_{align} = \frac{1}{B} \sum_{b=1}^{B} \sum_{i \neq j, i,j \in \{1,2,3\}} (\mathbf{r}_{b,i} \cdot \mathbf{r}_{b,j})^2,
\end{equation} 
where $\mathbf{r}_{b,i}$ is the i-th row vector of $\mathbf{M}$ for the $b$-th image in a batch of size $B$. 

\subsection{LUT Correction in Chromatic Aberration Space} Once the image is represented in CAS, we introduce our core innovation: a direction-aware correction mechanism that operates on the decoupled channels. This process is motivated by the physics of LCA; since the artifact is a spatial misalignment of color, its correction must be spatially informed. To this end, we employ a novel cross-channel guidance strategy, where the geometric properties of the fringe channel are used to steer the correction of the luminance channel.

This strategy is embodied in our \textbf{Direction-Aware 5D Look-Up Table (LUT)}, designed to adjust the luminance channel ($C_{lum}$) by incorporating spatial context to prevent such artifacts. Instead of being a simple color mapper, its lookup coordinates are designed to fuse spatial context with chromatic directional cues. 
For each pixel $p$, the five dimensions for LUT lookup include the luminance values of the center pixel $C_{lum}(p)$, the pixel to the left $C_{lum}(p_{left})$, and the pixel above $C_{lum}(p_{top})$, where non-symmetric, L-shaped sampling effectively captures edge information, along with the horizontal $\nabla_x C_{fringe}(p)$ (dx) and vertical $\nabla_y C_{fringe}(p)$ (dy) gradients of the fringe channel, enabling encode the direction and magnitude of the purple fringe at pixel. By conditioning the luminance correction on the fringe's gradient, the 5D-LUT learns to locally counteract the color channel misalignment at the heart of LCA, especially around high-frequency areas where fringing occurs, yielding the corrected luminance channel $C'_{lum}$.

Concurrently, the fringe channel ($C_{fringe}$) is refined by a simpler, learnable \textbf{1D LUT}. This LUT performs a global mapping to suppress the residual fringe intensity across the entire image, producing the corrected fringe channel $C'_{fringe}$. 
To ensure the learned transformations are smooth and avoid introducing new artifacts, we apply a \textbf{smoothness loss}, ${L}_{smooth}$, which penalizes large gradients between adjacent entries in the LUTs. It is defined as: 

\begin{equation} {L}_{smooth} = ||\nabla^2 \mathbf{L}_{1D}||_F^2 + \sum_{d=1}^{5} ||\nabla_d \mathbf{L}_{5D}||_F^2,
\end{equation} 
where $\nabla^2$ and $\nabla_d$ are the second-order and first-order finite difference operators, respectively. After correction, the corrected channels $C'_{lum}$ and $C'_{fringe}$, along with the original orthogonal channel $C_{ortho}$, form the final corrected CAS image. This image is then transformed back to the RGB domain via the analytical inverse of the transformation matrix, $\mathbf{M}^{-1}$, to yield the final output image $\mathbf{I}_{out}$: 
\begin{equation} \mathbf{I}_{out} = \mathbf{M}^{-1} \cdot \begin{pmatrix} C'_{lum} \\ C'_{fringe} \\ C_{ortho} \end{pmatrix}. \end{equation}

The fidelity of this output is governed by a comprehensive \textbf{reconstruction loss}, ${L}_{rec}$. This loss is a weighted sum of a standard L1 pixel loss (${L}_{L1}$) and our \textbf{YCbCr Perceptual Loss} (${L}_{p}$): 
\begin{equation} {L}_{rec} = \lambda_{L1}{L}_{L1} + \lambda_{p}{L}_{p}.
\end{equation} 
The ${L}_{p}$ term operates in the YCbCr color space. It computes a perceptual loss on the luma (Y) channel by measuring the L1 distance between deep features extracted by a pre-trained VGG19 network ($\phi$)~\cite{simonyan2014veryVGG}. Simultaneously, it computes a direct L1 loss on the chroma (Cb, Cr) channels. This hybrid approach ensures both structural/textural fidelity on the luminance and color accuracy on the chrominance. It is defined as: 
\begin{equation} 
    \begin{aligned} 
    {L}_{p} ={}& ||\phi(\mathbf{Y}_{out}) - \phi(\mathbf{Y}_{gt})||_1 \\ & + \lambda_{chroma}(||\mathbf{Cb}_{out} - \mathbf{Cb}_{gt}||_1 \\ & + ||\mathbf{Cr}_{out} - \mathbf{Cr}_{gt}||_1).
    \end{aligned} 
\end{equation} 

\subsection{Total Objective}
The entire DCA-LUT framework is optimized end-to-end by minimizing a composite objective function. As implemented in the provided code, this function is a weighted sum of the reconstruction and regularization loss components:
\begin{equation}
\label{eq:total_loss}
{L}_{total} = {L}_{rec} + \lambda_{1}{L}_{smooth} + \lambda_{2}{L}_{align}.
\end{equation}
This comprehensive objective function ensures that the network not only produces a high-fidelity reconstruction but also learns a physically plausible, stable, and regularized internal mechanism for purple fringing removal.

\begin{table*}[htbp]
    \centering
    \setlength{\tabcolsep}{2.5pt}
    \begin{tabular}{l c ccccc ccc}
        \toprule
        \multirow{2}{*}{\textbf{Method}} & \multirow{2}{*}{\textbf{Venue}} &
\multicolumn{5}{c}{\textbf{Quality Metrics}} &
\multicolumn{3}{c}{\textbf{Efficiency Metrics}} \\
        \cmidrule(lr){3-7} \cmidrule(lr){8-10}
        & & \textbf{PSNR} $\uparrow$ & \textbf{SSIM} $\uparrow$
& \textbf{LPIPS} $\downarrow$ & \textbf{ECAS} $\downarrow$ &
 \textbf{$\Delta$E} $\downarrow$ & \textbf{Params (M)} $\downarrow$
& \textbf{Time (s) 2K} $\downarrow$ & \textbf{Time (s) 4K}
$\downarrow$ \\
        \midrule
        SRLUT %\cite{jo2021practicalSRLUT}
        & CVPR'21 & 38.376 & 0.9624 & 0.0411 & 0.0632 & 0.9581
 & 0.15 & 0.0402 & 0.1768 \\
        SwinIR
        %\cite{liang2021swinir}
        & ICCV'21 & 36.486 & 0.7923 & 0.1232 & 0.1755 & 2.3823
 & 11.50 & OOM & OOM \\
        Restormer
        %\cite{zamir2022restormer}
        & CVPR'22 & 37.856 & 0.8896 & 0.0879 & 0.0662 & 2.1612
 & 26.11 & OOM & OOM \\
        DDRM %\cite{kawar2022denoisingDDRM}
        & NeurIPS'22 & \textbf{39.154} & \underline{0.9734} &
\underline{0.0317} & \underline{0.0411} & \underline{0.9581}
& 72.52 & OOM & OOM \\
        Lightendiffusion
        %\cite{Jiang_2024_ECCV}
        & ECCV'24 & 37.765 & 0.8724 & 0.0672 & 0.0715 & 0.9241
 & 27.83 & 0.9624 & OOM \\
        NILUT
        %\cite{conde2024nilut}
        & AAAI'24 & 38.445 & 0.9681 & 0.0416 & 0.0686 & 1.4218
 & \textbf{0.03} & 0.0549 & 0.2794 \\
        DnLUT
        %\cite{yang2025dnlut}
        & CVPR'25 & 38.902 & 0.9642 & 0.0627 & 0.0831 & 1.5526
 & 0.59 & 4.7505 & OOM \\
        BPAM
        %\cite{lou2025learninBPAM}
        & ICCV'25 & 38.165 & 0.9671 & 0.0588 & 0.1085 & 1.5755
 & 11.68 & 0.4851 & OOM \\
        \midrule
        \textbf{Ours} & - & \underline{39.052} & \textbf{0.9849}
& \textbf{0.0303} & \textbf{0.0438} & \textbf{0.6784} &
\underline{0.13} & \textbf{0.0405} & \textbf{0.1756} \\
        \bottomrule
    \end{tabular}
    \caption{Quantitative comparison with state-of-the-art methods on
the purple fringing benchmark. For PSNR, SSIM, higher is better
($\uparrow$). For LPIPS, ECAS, $\Delta$E, Parameters, and Inference
Time, lower is better ($\downarrow$). The best results are in
\textbf{bold}, and the second-best are \underline{underlined}. OOM
stands for ``Out of Memory", indicating that the method failed to
process the corresponding resolution due to insufficient memory.}
    \label{tab:sota_comparison_detailed}
\end{table*}

\section{Experiments} 
\subsection{Setup} 
% \noindent\textbf{Implementation Details.} 
Our framework is implemented using PyTorch and trained on a single NVIDIA RTX 3090 GPU. We use the Adam optimizer with a weight decay of $5 \times 10^{-4}$. The initial learning rate is set to $5 \times 10^{-5}$ and is decayed using a Cosine Annealing schedule over 150 epochs. The batch size is set to 16. For our model's architecture, the 5D LUT for luminance correction has a resolution of $9^5$, while the 1D LUT for fringe channel correction has a size of 1024. The entire training process is end-to-end. 

\subsection{Datasets} 
% \textbf{Datasets.} 
The absence of large-scale, paired datasets for purple fringing removal necessitates the creation of a synthetic dataset for robust training and quantitative evaluation. To this end, we introduce \textbf{PF-Synth}, a new synthetic dataset constructed using high-quality video frames from the DAVIS 2017 and 2019 datasets~\cite{Pont-Tuset_arXiv_2017, Caelles_arXiv_2019}, where frames from DAVIS 2017 are used for training, while those from DAVIS 2019 serve as the test and validation sets.
The synthesis pipeline, detailed in the Appendix, is designed to mimic the physical properties of Longitudinal Chromatic Aberration. Given a clean image, we first detect its high-contrast edges using a Canny edge detector. A random subset of these edge pixels is then selected to ensure the fringing appears sparsely, as it does in reality. A mask is generated by dilating these sparse edge points, which is then blurred to create smooth transitions. Finally, this mask is used to blend a predefined purple color with the original image, with randomized parameters for intensity, width, and sparsity to ensure diversity. For comprehensive evaluation, we also collected real-world purple fringe photos from photography forums to assess the model's practical performance in real scenarios, with various consumer-grade lenses exhibiting noticeable purple fringing.

\begin{figure*}[t]
    \centering
    \includegraphics[width=\textwidth, height=0.4\textheight]{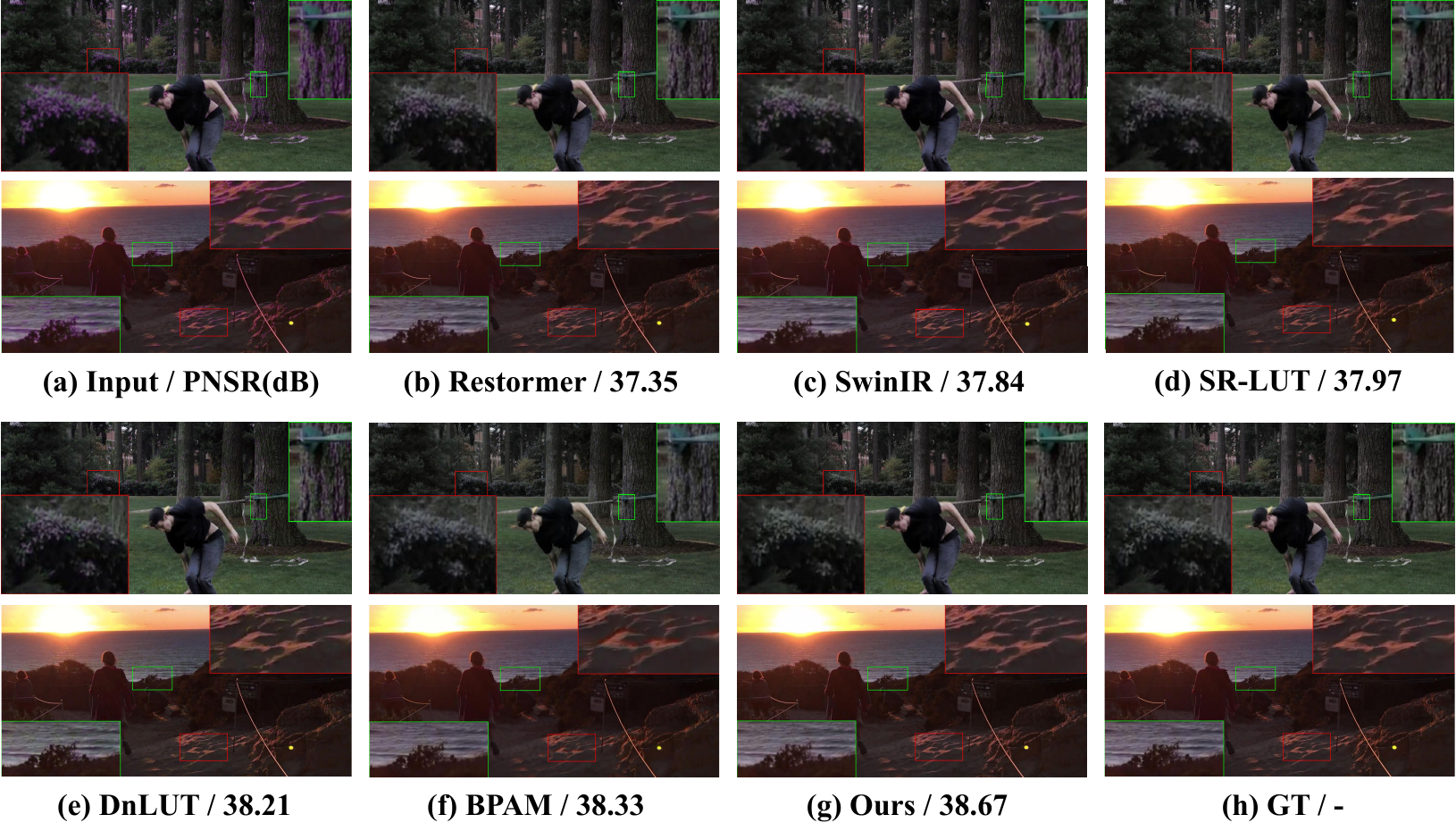}
    \caption{Visual comparison on our PF-Synth dataset. While competing methods often introduce secondary artifacts such as \textbf{darkening}, \textbf{color casts}, or \textbf{detail loss}, our approach is the only one that cleanly removes the purple fringe while faithfully preserving the original image content. (Please zoom in for the best view.)}
    \label{fig:quantative compare}
\end{figure*}

 \subsection{Evaluation Metrics} 
 \noindent\textbf{Standard Metrics.} 
 To assess the overall image restoration quality, we employ three widely-used metrics: Peak Signal-to-Noise Ratio (PSNR) and Structural Similarity Index (SSIM), which measure pixel-level fidelity and structural preservation, respectively. Additionally, we use the Learned Perceptual Image Patch Similarity (LPIPS), which better aligns with human perceptual judgment by comparing deep features. We also utilize the CIEDE2000 color difference metric \textbf{$\Delta$E} for quantifying the perceptual difference between two colors.
 %
 % For efficiency evaluation, we report the number of model parameters (Params) and the average inference time on a single NVIDIA RTX 4090 GPU for a 2K and 4K resolution image. 

 \noindent\textbf{Specialized Metric for Chromatic Aberration.} General metrics like PSNR and SSIM, while useful, are often insensitive to localized and color-specific artifacts such as purple fringing. A high PSNR score does not guarantee the complete removal of subtle color fringes along high-contrast edges. To address this limitation and provide a more accurate evaluation for this specific task, we propose a novel metric: the \textbf{Edge Chromatic Aberration Score (ECAS)}. The core idea of ECAS is to first identify high-contrast edge regions (where fringing occurs) using the ground-truth image, and then quantify the proportion of pixels within these regions that exhibit undesirable coloration in the corrected image. A lower ECAS score indicates a more effective suppression of chromatic aberration.
 Specifically, for a given corrected image $\mathbf{I}_{corr}$ and ground-truth $\mathbf{I}_{gt}$, the calculation involves three steps. First, an edge mask $\mathcal{M}_{edge}$ is generated from $\mathbf{I}_{gt}$ by applying a Sobel filter and thresholding the gradient magnitude at $\tau_{edge}$. This mask identifies all high-contrast regions of interest. Second, an aberration mask $\mathcal{M}_{aberration}$ is created from $\mathbf{I}_{corr}$ by converting it to the HSV color space and identifying pixels whose hue falls within predefined purple/green ranges and whose saturation exceeds a threshold $\tau_{sat}$. The final ECAS score is the ratio of the fringe pixels detected on edges to the total number of edge pixels:
\begin{equation}
\label{eq:ecas}
\text{ECAS} = \frac{\sum (\mathcal{M}_{edge} \odot \mathcal{M}_{aberration})}{\sum \mathcal{M}_{edge}},
\end{equation}
where $\odot$ denotes element-wise multiplication. A lower ECAS score indicates a more effective suppression of chromatic aberration, with a score of 0 being a perfect removal. This targeted metric serves as a crucial tool for evaluating model performance and validating our design choices.

 % \subsection{Comparison with State-of-the-Art Methods}
 \subsection{Quantitative and Qualitative Evaluation}
 We conduct a comprehensive comparison against a range of state-of-the-art (SOTA) methods, spanning different architectures and publication years. The competitors include LUT-based models (SRLUT~\cite{jo2021practicalSRLUT}, NILUT~\cite{conde2024nilut}, DNLUT~\cite{yang2025dnlut}, BPAM~\cite{lou2025learninBPAM}), powerful Transformer-based general image restorers (SwinIR~\cite{liang2021swinir}, Restormer~\cite{zamir2022restormer}), and recent diffusion-based models (Lightendiffusion~\cite{Jiang_2024_ECCV}, DDRM~\cite{kawar2022denoisingDDRM}). 
 
 \noindent\textbf{Quantitative Comparison.} Table~\ref{tab:sota_comparison_detailed} presents the quantitative results on our purple fringing benchmark. The results clearly demonstrate the superiority of our proposed method. Our model achieves state-of-the-art results in several key quality metrics, securing the best SSIM (0.9849), LPIPS (0.0303), and color accuracy with the lowest $\Delta E$ (0.6784). While the diffusion-based model DDRM achieves the highest PSNR and lowest ECAS, our method is a very close second in both categories, confirming its top-tier reconstruction fidelity and artifact removal capability. Notably, our ECAS score of \underline{0.0438} is significantly better than those of transformers like SwinIR and Restormer, underscoring the effectiveness of our specialized, physics-informed design.
 
% Most impressively, our model excels in efficiency. It is the second most lightweight model with only \underline{0.13M} parameters and achieves the fastest inference speed for both 2K (0.0405 ms) and 4K (0.1756 ms) images. This is in stark contrast to the large transformer and diffusion models (SwinIR, Restormer, DDRM), which run out of memory (OOM) on high-resolution inputs. Overall, our method strikes a superior balance between performance and efficiency, establishing itself as not only one of the most effective but also the most practical solution for real-world purple fringing removal. 

 \noindent\textbf{Qualitative Comparison.} As shown in the visual comparisons in Fig.~\ref{fig:quantative compare}, our method produces qualitatively superior results. General-purpose models like SwinIR, Restormer, despite their power, tend to either leave residual purple fringe or over-smooth fine details along the edges. 
 Specialized LUT-based methods also exhibit distinct drawbacks. Both SR-LUT and DnLUT manage to suppress the purple color but do so at the cost of creating unnatural dark outlines that degrade local contrast. BPAM fails more catastrophically by replacing the purple fringe with severe yellowish and greenish color artifacts, demonstrating a fundamental lack of robustness.
 In contrast, our method cleanly removes the chromatic aberration while faithfully preserving the underlying scene. Our results are visually the most plausible and closest to the ground truth, proving that our approach can precisely eliminate the artifact without causing collateral damage to the image content.

\begin{table}[htbp]
    \centering
    \setlength{\tabcolsep}{8pt}  % 进一步减小列间距
    \begin{tabular}{l c c}
        \toprule
        \textbf{Configuration} & \textbf{PSNR} $\uparrow$ & \textbf{ECAS} $\downarrow$ \\
        \midrule
        % --- Channel Ablation ---
        Correct Green Channel        & 38.104 &  0.0957     \\
        Correct Luma Channel (Ours)  & \textbf{39.052} & \textbf{0.0438} \\
        \midrule
        % --- LUT Design Ablation ---
        Symmetric LUT            & 37.552 & 0.1512 \\
        + Asymmetric Sampling    & 38.827 & 0.0865 \\
        + Gradient Guide (Ours)  & \textbf{39.052} & \textbf{0.0438} \\
        \bottomrule
    \end{tabular}
    \caption{Comprehensive ablation studies validating our key design choices.}
    \label{tab:ablation_unified}
\end{table}

\subsection{Ablation Studies} 
% \noindent\textbf{Effectiveness of Chromatic-Aware Coordinate Transformation.} 
To validate the critical components of our DCA-LUT framework, we conduct a series of ablation experiments. We systematically analyze two key design choices: the selection of the luminance channel for correction, and the architecture of our Direction-Aware 5D LUT.

% \noindent\textbf{Effectiveness of the CA-CT Module.}
% A cornerstone of our method is the CA-CT module, which learns to transform the input image into a space where the purple fringe is isolated. We first compare this approach to traditional methods that rely on handcrafted features to detect fringe regions. Heuristic-based methods can often misidentify legitimate purple objects as artifacts or fail to capture the fringe completely. In contrast, our CA-CT module, by learning the transformation from data, can precisely decouple the purple fringe into its own channel. This is because our method is motivated by the physical root of LCA: it learns to untangle the spatially misaligned RGB channels into a more meaningful representation, ensuring that only the true artifact is isolated for correction.

\begin{figure}[h]
\centering
\includegraphics[width=\columnwidth]{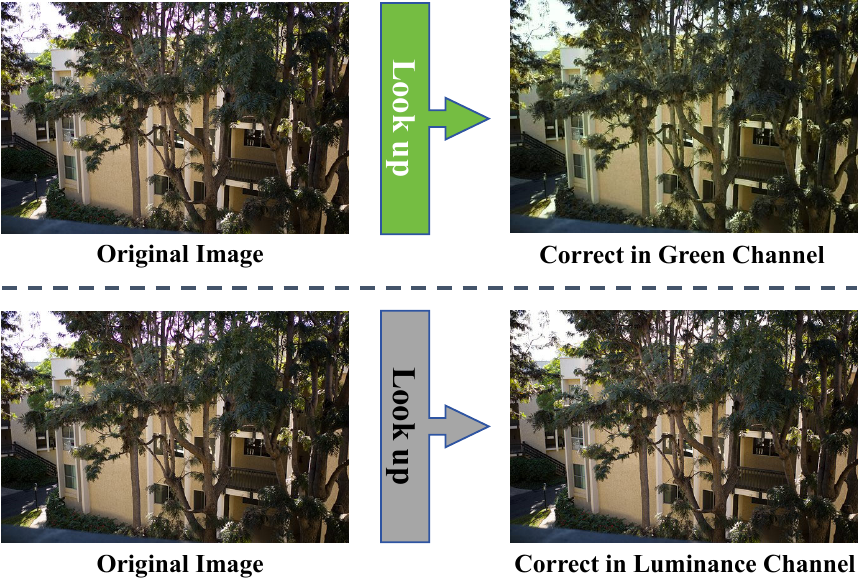}
\caption{Comparison of correction strategies. Green-channel correction (top) introduces a noticeable green tint, while our luminance-based approach (bottom) preserves natural color fidelity.}
\label{fig:green_vs_lum}
% \vspace{-0.2cm}
\end{figure}

% \noindent\textbf{Why Correct Luminance and Not the Green Channel?}
\noindent\textbf{Advantages of Luminance-Based Correction over Green Reference.} Given that LCA is a misalignment of the red and blue channels relative to the green channel, a plausible alternative is to use the green channel as the reference for correction. However, this approach has a significant drawback. As illustrated in Fig.~\ref{fig:green_vs_lum}, directly correcting the image based on the green channel introduces a noticeable greenish color cast across the entire image, altering the global color balance. By using our CA-CT module to generate a dedicated luminance channel ($C_{lum}$), our framework can perform corrections in a color-neutral space. This ensures that the artifact is removed without introducing undesirable color shifts, leading to a more natural and physically accurate result.

\begin{figure}[h]
\centering
\includegraphics[width=\columnwidth]{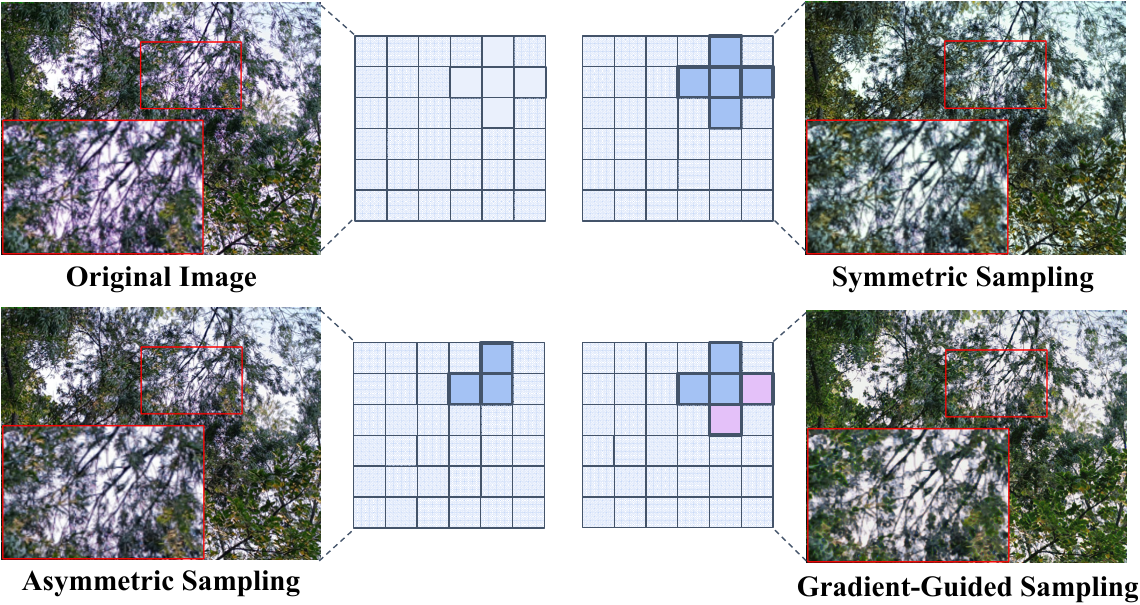}
\caption{Visual ablation of the 5D LUT design. The gradient-guided approach (bottom right) most effectively removes the purple fringe without introducing artifacts, demonstrating the superiority of the direction-aware design.}
\label{fig:lut_ablation_visual}
% \vspace{-0.4cm}
\end{figure}

\noindent\textbf{Effectiveness of the Direction-Aware 5D LUT.} We validate the design of our core correction module, the Direction-Aware 5D LUT. We compare our design against two simpler variants to demonstrate the importance of both its asymmetric structure and the cross-channel gradient guidance.
The qualitative results in Fig.~\ref{fig:lut_ablation_visual} and quantitative metrics in Table \ref{tab:ablation_unified} clearly show the progression. The Symmetric LUT struggles to precisely target the fringe, often causing blurring or incomplete removal because it averages information across the edge. The Asymmetric (L-shaped) LUT performs better by capturing the edge-like nature of the artifact more effectively. However, our final Gradient-Guided LUT achieves the best results. By explicitly using the fringe's direction and magnitude as a guide, it performs the most precise correction, completely removing the artifact while preserving the sharpness and integrity of the underlying edge. 

% This confirms that our direction-aware, cross-channel guidance strategy is critical for achieving strong performance.

\section{Discussion}

\noindent\textbf{Architectural Rationale: 5D LUT for Structural Repair, 1D LUT for Global Color.} Purple fringing is edge-localized and structure-dependent. Our 5D LUT, in the Chromatic Aberration space with luminance/fringe cues and edge geometry $(|\nabla|)$, performs direction-aware local corrections on luminance, removing fringes while preserving detail. Global color fidelity is direction-invariant; a lightweight 1D LUT serves as the final tone curve, is easily regularized (with monotonicity constraints), and avoids direction-dependent casts at negligible cost. This separation lets 5D fix the structure and 1D enforce global color consistency. As shown in Table~\ref{tab:sota_comparison_detailed}, our architecture delivers state-of-the-art performance with minimal computational cost.
%
% Ablations show: 5D-only lowers ECAS but raises $\Delta E$/casts; 1D-only preserves color but leaves edge residuals; 5D+1D achieves the lowest ECAS and lower $\Delta E$ with the same runtime.

\begin{figure}[h]
\centering
\includegraphics[width=\columnwidth]{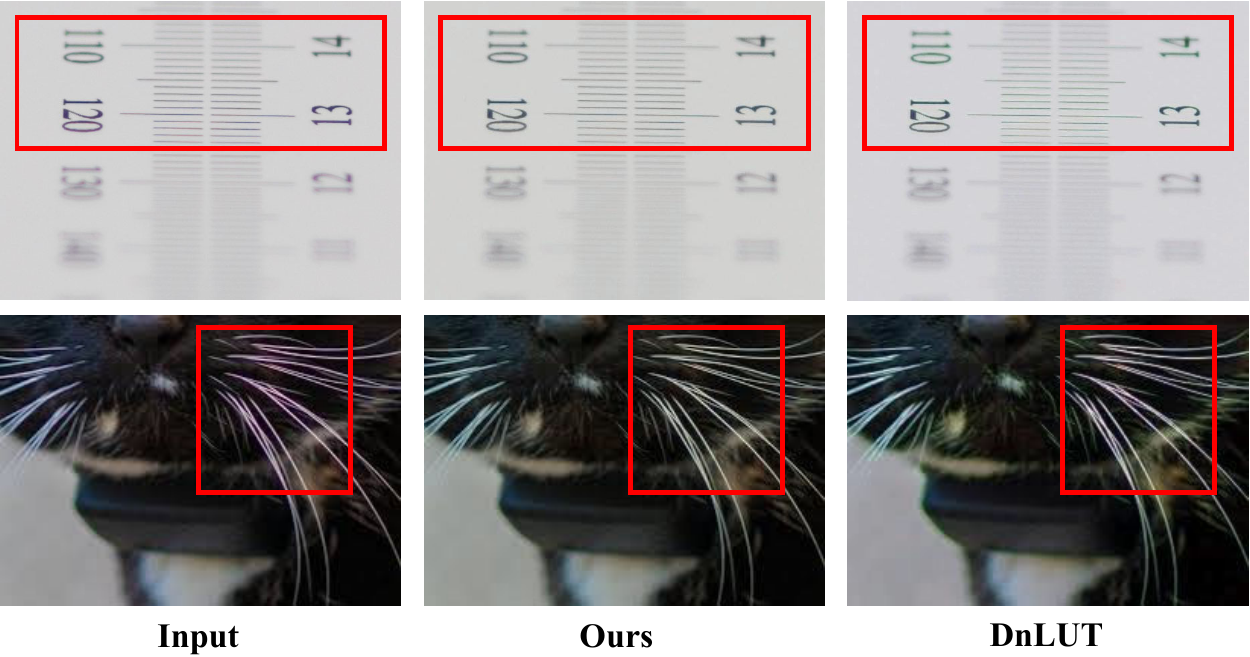}
\caption{Visual comparison on real-world images. The competing method introduces a noticeable \textbf{greenish cast} and leaves \textbf{residual purple fringes}, while our method removes the fringe cleanly and preserves natural brightness and texture.}
\label{fig:realword}
% \vspace{-0.6cm}
\end{figure}

\noindent\textbf{Real-world Purple Fringing Removal.} We validate our model on challenging real-world images, which often exhibit more complex fringing than synthetic data. As shown in Fig.~\ref{fig:realword}, both our method and the competing specialized method DnLUT can suppress the purple hue. However, DnLUT introduces a significant darkening artifact, creating an unnatural dark halo around the high-contrast edges of the ruler's markings and the cat's whiskers. In contrast, our method cleanly eliminates the fringe while faithfully preserving the original brightness and texture of the scene. This demonstrates our model's superior robustness and fidelity in practical scenarios, proving it can precisely target the artifact without causing collateral damage.

\noindent\textbf{Model Complexity and Efficiency.} Our model excels in efficiency. For evaluation, we report the number of parameters (Params) and the average inference time on a single NVIDIA RTX 3090 GPU for 2K and 4K resolution images. As shown in Table~\ref{tab:sota_comparison_detailed}, our method is the second most lightweight model, with only 0.13M parameters, and achieves the fastest inference speeds—0.0405 s for 2K and 0.1756 s for 4K. This stands in stark contrast to large transformer and diffusion-based models (e.g., SwinIR, Restormer, DDRM), which fail to process high-resolution inputs due to memory constraints. Overall, our method achieves an excellent trade-off between performance and efficiency.

% making it not only highly effective but also practically deployable for real-world purple fringing removal.

% \noindent\textbf{Complexity of the Model.} Our model excels in efficiency. As shown in Table 1, It is the second most lightweight model with only 0.13M parameters and achieves the fastest inference speed for both 2K (0.0405 ms) and 4K (0.1756 ms)  
% Look up Look up images. This is in stark contrast to the large transformer and diffusion models (SwinIR, Restormer, DDRM), which run out of memory on high-resolution inputs. Overall, our method strikes a superior balance between performance and
% efficiency, establishing itself as not only one of the most effective but also the most practical solution for real-world purple fringing removal.

% \noindent\textbf{Inference on Mobile Devices.} To validate the real-world viability of our lightweight design, we deployed DCA-LUT on a mobile device equipped with . Our model demonstrated remarkable performance, processing a standard 12-megapixel image in approximately 150ms.

\section{Conclusion}
In this paper, we presented DCA-LUT, a novel and efficient framework for purple fringing removal. Motivated by the physics of Longitudinal Chromatic Aberration, our method first learns a transformation to isolate the fringe artifact, and then uses a Direction-Aware 5D LUT that leverages the fringe's geometric gradient to steer the luminance restoration.
Extensive experiments show that DCA-LUT achieves outstanding results, cleanly removing artifacts without the side effects seen in competing methods, all while being orders of magnitude faster and more memory-efficient.  
Moreover, we validated our model's robustness through real-world tests across diverse scenarios and deployed it on mobile devices. Our work not only provides a practical solution to a persistent optical problem but also demonstrates a new paradigm for using learnable LUTs as precise, spatially-aware correction operators.

\section{Acknowledgments}
This work was supported in part by the National Natural Science Foundation of China under Grant 62172265, in part by the Shandong Provincial Natural Science Foundation under Grant ZR2025MS1025.

\bibliography{main}

\clearpage

\appendix

\twocolumn[
  \begin{center}
    {\LARGE \textbf{Supplementary Material}} % 这里是大标题，\Huge 控制大小
    \vspace{1cm}
  \end{center}
]
\section{Overview of Appendix}

This appendix provides supplementary materials to support the main paper. We detail the key components of our experimental setup to ensure the clarity of our work. The contents are organized as follows:
\begin{itemize}
    \item \textbf{PF-Synth Dataset Synthesis:} We provide a detailed description and the algorithmic pipeline for generating our synthetic purple fringing dataset, PF-Synth.
    \item \textbf{Real-world Image Processing:} We explain the sourcing and utilization of real-world images for the qualitative evaluation of our method, showcasing its robustness against authentic, complex fringing artifacts found in everyday photography.
\end{itemize}

\section{PF-Synth Dataset Synthesis}

To facilitate robust, data-driven training and standardized evaluation, we constructed a large-scale synthetic dataset, PF-Synth. The synthesis pipeline is meticulously designed to mimic the physical properties of Longitudinal Chromatic Aberration (LCA).

The process begins with a high-quality, artifact-free image. We first identify high-contrast edges using a Canny edge detector, as these are the regions where fringing artifacts are most prominent in reality. To ensure the sparse appearance characteristic of real-world fringing, we randomly select a subset of these edge pixels. A mask is then generated by dilating these sparse points, which is subsequently blurred using a Gaussian filter to create smooth, natural transitions between the fringe and the original image content. Finally, this soft mask is used to blend a predefined purple color with the original image pixels. The entire process incorporates randomized parameters for fringe intensity, width, and sparsity to generate a diverse and comprehensive dataset. The detailed procedure is outlined in Algorithm \ref{alg:synthesis}.

% --- 算法伪代码 ---
\begin{algorithm}[tb]
\caption{Purple Fringing Synthesis Pipeline} \label{alg:synthesis}
\textbf{Input}: Clean image $\mathbf{I}_{clean}$, intensity range $[\alpha_{min}, \alpha_{max}]$, width range $[w_{min}, w_{max}]$, sparse ratio range $[\rho_{min}, \rho_{max}]$ \\
\textbf{Output}: Image with synthetic purple fringe $\mathbf{I}_{fringe}$
\begin{algorithmic}[1]
\STATE $\alpha \leftarrow \text{RandomUniform}(\alpha_{min}, \alpha_{max})$
\STATE $w \leftarrow \text{RandomInt}(w_{min}, w_{max})$
\STATE $\rho \leftarrow \text{RandomUniform}(\rho_{min}, \rho_{max})$
\STATE $\mathbf{I}_{gray} \leftarrow \text{ConvertToGrayscale}(\mathbf{I}_{clean})$
\STATE $\mathbf{E} \leftarrow \text{CannyEdgeDetection}(\mathbf{I}_{gray})$
\STATE $\mathcal{P}_{edge} \leftarrow \text{GetEdgePixelCoordinates}(\mathbf{E})$
\IF{$|\mathcal{P}_{edge}| > 0$}
\STATE $\mathcal{P}_{sparse} \leftarrow \text{RandomSample}(\mathcal{P}_{edge}, \rho \cdot |\mathcal{P}_{edge}|)$
\STATE $\mathbf{M}_{sparse} \leftarrow \text{CreateMaskFromPoints}(\mathcal{P}_{sparse}, w)$
\STATE $\mathbf{M}_{blur} \leftarrow \text{GaussianBlur}(\mathbf{M}_{sparse})$
\STATE $\mathbf{C}_{purple} \leftarrow [0.6, 0.0, 0.8]$ \COMMENT{Define purple color in RGB}
\STATE $\mathbf{I}_{fringe} \leftarrow (1 - \alpha \cdot \mathbf{M}_{blur}) \odot \mathbf{I}_{clean} + \alpha \cdot \mathbf{M}_{blur} \odot \mathbf{C}_{purple}$
\ELSE
\STATE $\mathbf{I}_{fringe} \leftarrow \mathbf{I}_{clean}$
\ENDIF
\STATE \textbf{return} $\text{Clip}(\mathbf{I}_{fringe}, 0, 1)$
\end{algorithmic}
\end{algorithm}
% ---------------------------------

\section{Real-world Image Processing}

While synthetic data is essential for quantitative analysis and training, evaluating performance on authentic, real-world images is crucial for assessing a model's generalization capabilities. To this end, we conducted a comprehensive qualitative evaluation using a collection of real-world photographs exhibiting noticeable purple fringing.

\begin{figure*}[h]
    \centering
    \includegraphics[width=\textwidth]
    {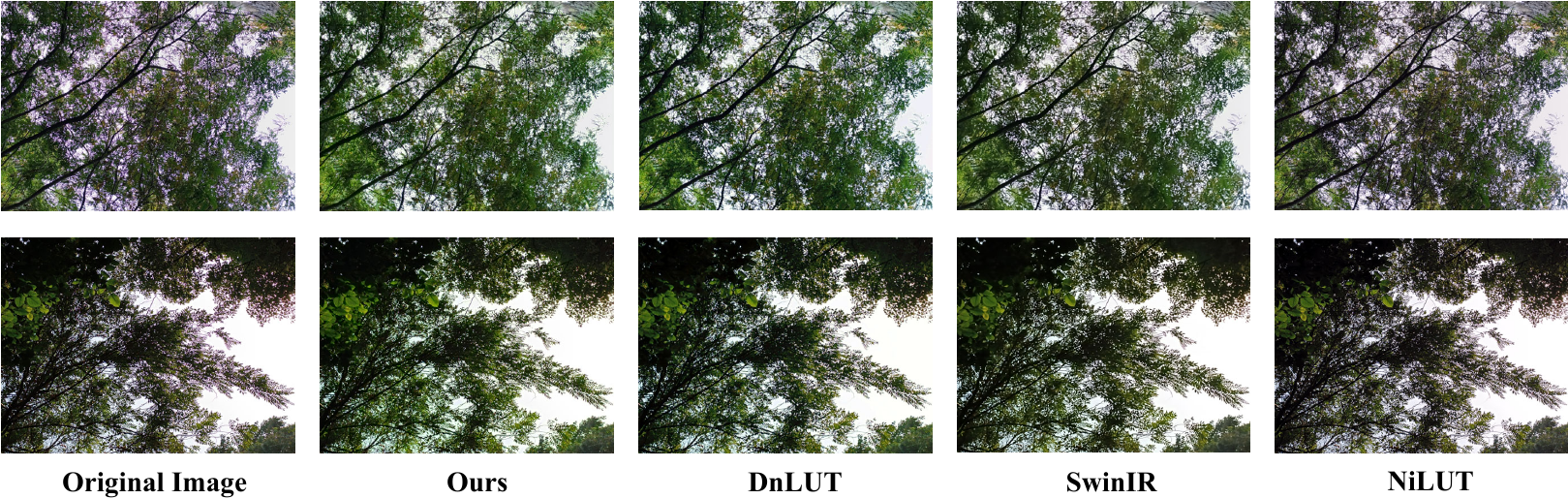}
    \caption{Qualitative comparison on real-world images. Our method successfully removes purple fringing from complex tree branches while preserving natural color and contrast. Competing methods either leave residual artifacts (SwinIR, NILUT) or introduce subtle darkening on the edges (DnLUT).}
    \label{fig:real}
\end{figure*}

These images were sourced from public photography forums, where photographers often share images captured with various consumer-grade lenses prone to significant chromatic aberration. This collection features a wide variety of scenes, lighting conditions, and fringing patterns, posing a challenging test for any correction algorithm. These images were not used for training but served exclusively as a testbed for qualitative comparison against other state-of-the-art methods. As shown in Fig.~\ref{fig:real}, our method demonstrates superior performance in complex real-world scenes. For instance, in the high-contrast areas of tree branches against the bright sky, competing methods like SwinIR and NILUT fail to completely eliminate the purple fringing, leaving distracting residual artifacts. While DnLUT is effective at suppression, it tends to slightly darken the edges, affecting the local contrast. In contrast, our model cleanly removes the aberration while faithfully preserving the natural color and brightness of the scene.

However, we also analyze the model's limitations in extremely challenging cases. As illustrated in the failure case in Fig.~\ref{fig:failure}, when faced with severe chromatic aberration on fine, complex structures like animal fur, our model does not achieve a perfect restoration. 
Nevertheless, it degrades more gracefully than other methods. Competing approaches like NILUT and BPAM not only fail to correct the initial artifact but also introduce severe, unnatural yellowish color casts. Our method, while imperfect, avoids introducing such distracting secondary artifacts, proving its robustness even at the edge of its operational capabilities.
\begin{center}
    \includegraphics[width=\columnwidth]{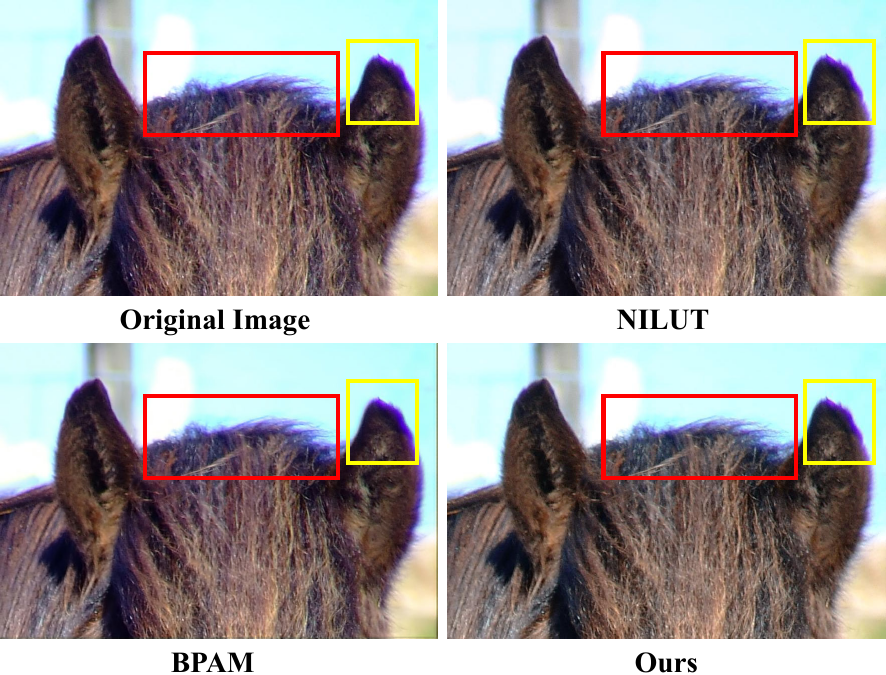}
    \captionof{figure}{A failure case analysis on an image with extreme chromatic aberration on fine fur. While our model does not perfectly remove the fringe, it avoids introducing the severe secondary color artifacts produced by competing methods like NILUT and BPAM, demonstrating more graceful degradation.}
    \label{fig:failure}
\end{center}

\end{document}